\RequirePackage{fix-cm}
\documentclass{article} 
\usepackage{iclr2027_conference,times}

\usepackage{amsmath,amsfonts,bm}

\def\eqref#1{equation~\ref{#1}}

\def\1{\bm{1}}

\DeclareMathAlphabet{\mathsfit}{\encodingdefault}{\sfdefault}{m}{sl}
\SetMathAlphabet{\mathsfit}{bold}{\encodingdefault}{\sfdefault}{bx}{n}

\usepackage{graphicx}
\usepackage{booktabs}
\usepackage{xcolor}
\usepackage{colortbl}
\definecolor{benchblue}{HTML}{DCECF5}
\definecolor{benchpeach}{HTML}{FBEBDD}
\definecolor{benchrose}{HTML}{F5E1E6}
\definecolor{benchgreen}{HTML}{E4F0E7}
\definecolor{benchgray}{HTML}{ECEFF2}
\usepackage{float}
\usepackage{capt-of}
\usepackage{tikz}
\usepackage{hyperref}
\hypersetup{hidelinks,
pdftitle={UMI-WM: Interactive World Modeling for Bimanual Manipulation from Wrist Cameras},
pdfauthor={Jingxiang Guo, Haoyu Zhao, Xidong Zhang, Yuhang Zheng, Hao Wu, Chen Gao, Jianshu Zhou, Shuicheng Yan, Weihao Yuan}}
\usepackage{xurl}
\usepackage{etoolbox}

\makeatletter
\patchcmd{\@maketitle}{\vskip 0.3in minus 0.1in}{\vskip 0.04in}{}{\PackageError{umi-arxiv}{Title spacing patch failed}{Check the title definition.}}
\makeatother

\title{\raggedright UMI-WM: Interactive World Modeling for Bimanual Manipulation from Wrist Cameras}

\author{\normalfont\begin{minipage}{\dimexpr\textwidth-2\tabcolsep\relax}
\centering
Jingxiang Guo$^{1}$\quad Haoyu Zhao$^{1}$\quad Xidong Zhang$^{2}$\quad Yuhang Zheng$^{1}$\quad Hao Wu$^{1}$\\
Chen Gao$^{1,\dagger}$\quad Jianshu Zhou$^{1,\dagger}$\quad Shuicheng Yan$^{1,\dagger}$\quad Weihao Yuan$^{3,\dagger}$\\[1pt]
{\small $^{1}$National University of Singapore\\
$^{2}$Harbin Institute of Technology\quad $^{3}$Nanjing University}\\[1pt]
{\footnotesize \href{mailto:jingxiangguo@u.nus.edu}{\texttt{jingxiangguo@u.nus.edu}}\quad $^{\dagger}$Corresponding authors}
\end{minipage}}
\iclrfinalcopy

\begin{document}
\raggedbottom
\setlength{\intextsep}{4pt plus 1pt minus 1pt}

\maketitle
\fancyhead{}
\renewcommand{\headrulewidth}{0pt}

\begin{figure}[H]
\setlength{\abovecaptionskip}{4pt}
\centering
\includegraphics[width=\linewidth]{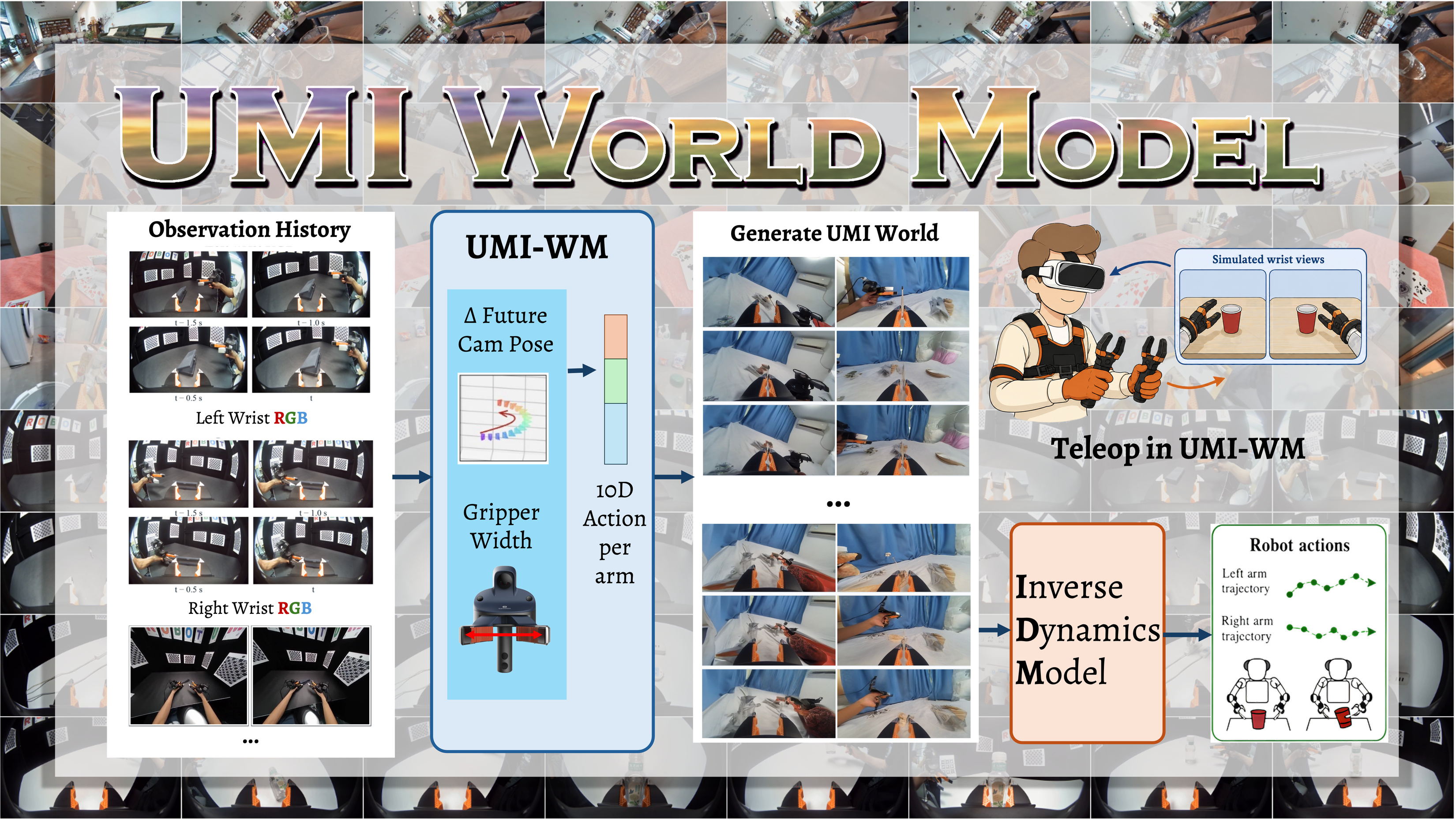}
\caption{UMI-WM jointly generates both wrist-camera views conditioned on bimanual actions, and an inverse dynamics model (IDM) recovers actions from the generated videos for evaluation and robot execution. More results and demos are on our \href{https://www.lv-lab.org/umi-wm/}{\textcolor{blue}{project page}}.}
\label{fig:teaser}
\end{figure}

\begin{abstract}
\looseness=-1
Universal Manipulation Interface (UMI) pairs wrist-camera videos with camera trajectories and gripper openings, providing scalable, robot-free data for learning how bimanual actions change a shared scene. Existing interactive world models, however, typically rely on text or single-camera control and struggle to follow each arm's actions while maintaining consistency between the two wrist views. We introduce UMI-WM, an action-conditioned bimanual world model that jointly generates both wrist views from observation history, per-arm relative camera poses, and gripper widths, with cross-view consistency, action following, and visual fidelity. UMI-WM injects each arm's actions into its own wrist stream, while sparse hub tokens exchange scene information between the two streams. Camera-aware geometry conditioning and camera-guided noise further support per-view motion control and cross-view geometric exchange. We adapt the model for block-causal generation and distill the 30-step Teacher into a Student with four denoising steps. We train UMI-WM on self-collected open-world and in-studio UMI recordings combined with public manipulation data. To evaluate action following, we use an inverse dynamics model (IDM) to recover relative camera poses and gripper widths from generated videos, and introduce Bimanual Action Fidelity (BAF) to measure how closely these recovered actions match the conditioning actions. UMI-WM achieves the best results on all five reference-video fidelity metrics and the best or competitive performance on VBench and WorldArena. The Teacher and Student outperform all compared baselines in action following, while the four-step Student generates paired wrist videos $6.9\times$ faster than the Teacher.
\end{abstract}

\newcommand{\draftnote}[1]{\par\noindent{\small\color{gray}\textit{Draft outline: #1}}\par}

\vspace{-6pt}
\section{Introduction}
\label{sec:introduction}
Recent interactive world models, including HY-WorldPlay~\citep{sun2025worldplay}, LingBot-World~\citep{robbyant2026lingbotworld}, Wonder~\citep{xu2026wonder}, and Matrix-Game 3.5~\citep{qian2026matrixgame35}, generate videos conditioned on camera poses, allowing users to control viewpoint motion through a scene. Extending this capability to bimanual manipulation requires modeling not only the motion of two independently controlled viewpoints but also how each end-effector's actions affect the shared environment. The Universal Manipulation Interface (UMI)~\citep{chi2024universalmanipulationinterfaceinthewild} provides a natural interface for this extension. Its handheld grippers collect wrist-camera videos, relative end-effector trajectories, and gripper widths without requiring a robot. Because each wrist camera is rigidly mounted to its gripper, the relative camera poses constitute the end-effector pose component of the action, while a single additional dimension specifies the gripper opening. Together, these controls provide a compact bimanual action representation\linebreak
for jointly predicting both future wrist-camera views.

Extending interactive world models to bimanual manipulation raises three consistency requirements. First, cross-view consistency: the two independently controlled wrist views must show the same shared scene, so that changes caused by either hand appear in both. Second, spatial consistency within each view: the persistent gripper foreground must stay stable while the scene moves, whereas existing models let it drift. Third, action consistency: the generated viewpoint motion and gripper opening must follow the commanded per-arm camera pose and width, including small pose changes.

To address these challenges, we introduce \textbf{UMI-WM}, an interactive world model that jointly predicts both wrist videos conditioned on bimanual actions (Fig.~\ref{fig:teaser}). UMI-WM injects each arm's actions into its own wrist stream, exchanges shared-scene context between the two streams through hub tokens, and adds camera-aware geometry conditioning, so that each wrist's motion and the scene changes caused by either hand are modeled together; causal generation and distilling the 30-step Teacher into a four-step Student keep the rollout efficient. Training uses a unified data processing pipeline that integrates self-collected UMI demonstrations with public robot data through a shared camera-centric action representation. Finally, an inverse dynamics model (IDM) recovers relative camera poses and gripper widths from generated videos, and Bimanual Action Fidelity (BAF) measures how closely these recovered actions agree with the conditioning actions.

Our contributions are then summarized as follows:
\begin{itemize}
\item \textbf{UMI-WM.} We develop an interactive world model that jointly generates both wrist-camera views conditioned on per-arm camera poses and gripper widths, combining cross-view interaction and camera-aware conditioning with few-step causal generation.
\item \textbf{Data pipeline and action-level evaluation.} We construct a three-tier data pyramid of public robot data, in-the-wild UMI recordings, and in-studio UMI recordings with a unified processing pipeline, and develop an IDM that recovers relative camera poses and gripper widths from video to compute Bimanual Action Fidelity (BAF).
\item \textbf{Results.} On 320 held-out UMI episodes, UMI-WM leads all five reference-video fidelity metrics and is best or competitive on VBench and WorldArena. The Teacher and Student reach BAF scores of 76.36 and 71.55, respectively, outperforming all compared baselines, while the four-step Student achieves a $6.9\times$ speedup over the 30-step Teacher.
\end{itemize}

\vspace{-6pt}
\section{Related Work}
\label{sec:related-work}

\vspace{-4pt}
\subsection{UMI Gripper and UMI Datasets}
UMI~\citep{chi2024universalmanipulationinterfaceinthewild} pairs wrist-camera videos, SLAM-reconstructed camera trajectories, and gripper widths for portable demonstration collection. Variants include FastUMI~\citep{zhaxizhuoma2025fastumi}, which simplifies hardware integration and pose tracking; DexUMI~\citep{xu2025dexumi}, which supports dexterous hands through a wearable exoskeleton; and MV-UMI~\citep{rayyan2025mvumi}, which adds third-person observations for broader scene context. Public datasets such as FastUMI-100K~\citep{liu2025fastumi100k}, RealOmni-Open~\citep{genrobot2025realomni}, and Daimon-Infinity~\citep{daimon2026infinity} broaden the coverage of manipulation tasks and sensing modalities. We combine public manipulation data with self-collected UMI recordings to learn a bimanual world model conditioned on relative wrist-camera poses and gripper widths.

\begin{figure}[H]
\centering
\setlength{\abovecaptionskip}{4pt}
\includegraphics[width=\linewidth]{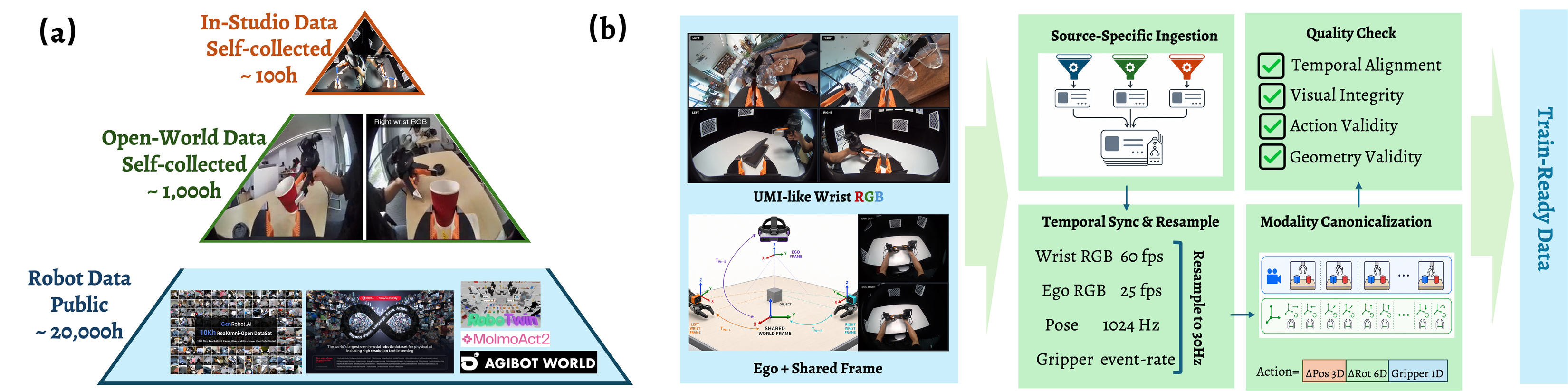}
\caption{\textbf{UMI data and processing pipeline.} (a) A three-tier data pyramid combines public robot data with open-world and in-studio UMI recordings. (b) The pipeline performs temporal alignment, action canonicalization, and quality checks across data sources.}
\label{fig:data-pipeline}
\vspace{-4pt}
\end{figure}

\vspace{-4pt}
\subsection{Camera-Controlled Interactive World Models}
Camera-controlled world models generate observations along camera trajectories. CameraCtrl~\citep{he2024cameractrl}, CamI2V~\citep{zheng2024cami2v}, and ReCamMaster~\citep{bai2025recammaster} introduce camera-pose conditioning into video generation, while HY-WorldPlay~\citep{sun2025worldplay} and LingBot-World~\citep{robbyant2026lingbotworld} support interactive exploration with persistent scene context. Extending this setting to multiple viewpoints requires modeling how independently controlled views evolve within a shared environment, as explored by SynCamMaster~\citep{bai2025syncammaster}, Gamma-World~\citep{liu2026gammaworld}, and Khora~\citep{zhao2026khora}. UMI-WM extends camera control to bimanual manipulation, where gripper actions change both wrist views and their shared scene.

\vspace{-4pt}
\subsection{World Models for Robotic Manipulation}
\textbf{World models as planners.} World models generate visual plans and then translate them into robot actions. UniPi~\citep{du2023unipi}, RoboDreamer~\citep{robodreamer2024}, and GE-Act 2.0~\citep{agibot2026geact2} couple visual planners with IDMs to recover actions. Large Video Planner~\citep{chen2025largevideoplanner} similarly extracts actions from visual plans, while Goal-VLA~\citep{chen2025goalvla} derives target object poses from generated goal images. UMI-WM also uses an IDM to recover actions from video and defines BAF by comparing them with the conditioning actions.

\vspace{-3pt}
\textbf{World models as simulators.} Genie Envisioner~\citep{liao2025genieenvisioner}, GE-Sim 2.0~\citep{qiu2026gesim2}, Interactive World Simulator~\citep{wang2026interactive}, and Ctrl-World~\citep{guo2026ctrlworld} generate visual responses to robot actions for policy evaluation, while RynnWorld-Teleop~\citep{zhao2026rynnworldteleop} enables teleoperation within a generative world model. UMI-WM serves as an interactive simulator for bimanual manipulation, generating wrist-camera videos conditioned on per-arm relative camera poses and gripper widths. Appendix~\ref{app:related-work} reviews further related work.

\vspace{-6pt}
\section{UMI Data and Processing Pipeline}
\label{sec:data}

We organize public robot data and self-collected UMI recordings into a three-tier pyramid with a unified processing pipeline for world model training (Fig.~\ref{fig:data-pipeline}).

\vspace{-4pt}
\subsection{UMI Data}
\label{sec:umi-data}
\textbf{Public manipulation data.} Public UMI datasets, including RealOmni-Open and Daimon-Infinity, provide UMI recordings aligned to a common action representation. MolmoAct2~\citep{fang2026molmoact2} and AgiBot World 2026~\citep{agibotworld2026} contribute robot wrist-camera videos with actions converted to the same representation. RoboTwin~\citep{chen2025robotwin2} and RoboLab~\citep{yang2026robolab} provide synthetic wrist-camera videos with relative camera poses and gripper widths.

\vspace{-3pt}
\textbf{Open-world UMI recordings.} We collect open-world bimanual demonstrations with paired DM-DataClaw devices (Appendix~\ref{app:umi-collection}, Fig.~\ref{fig:dataclaw-hardware}). DM-DataClaw is a tweezer-style UMI gripper; each device records wrist-camera video, tracks camera pose through SLAM, and measures gripper width. The two camera trajectories remain in separate local frames.

\begin{figure}[t]
\centering
\setlength{\abovecaptionskip}{4pt}
\includegraphics[width=\linewidth]{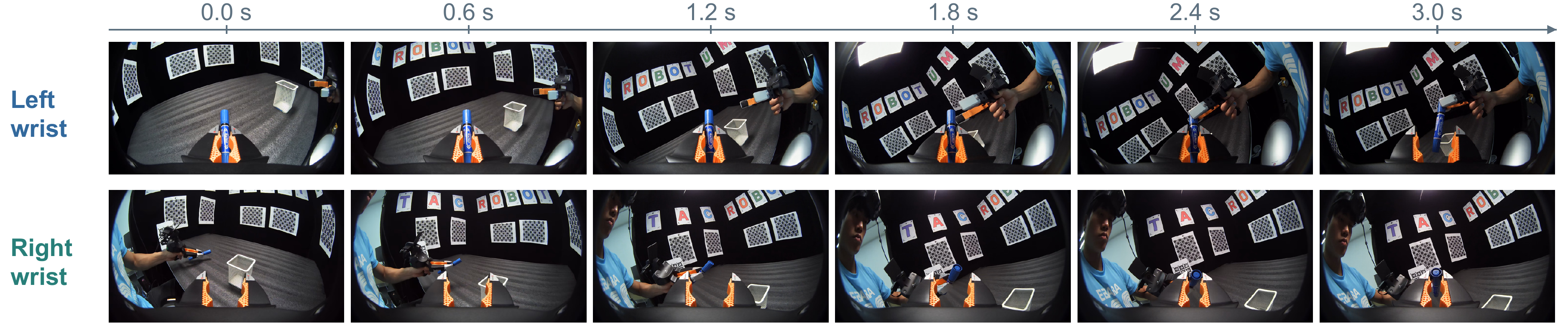}
\caption{\textbf{In-studio bimanual recording.} Synchronized left- and right-wrist views of a marker handover, sampled every 0.6\,s along a shared timeline. AprilGrid targets support camera registration.}
\label{fig:dataclaw}
\vspace{-4pt}
\end{figure}

\vspace{-3pt}
\textbf{In-studio UMI recordings.} Our studio aligns both wrist trajectories in a shared world frame. AprilGrid geometry~\citep{furgale2013kalibr} and calibrated intrinsics yield synchronized inter-wrist poses for cross-view geometric supervision (Appendix~\ref{app:umi-collection}, Fig.~\ref{fig:umi-collection-booth}). The public RealOmni-Open and Daimon-Infinity recordings do not provide this shared-frame registration. We record synchronized wrist and stereo egocentric RGB videos; UMI-WM uses only the two wrist views with camera poses and gripper widths as conditioning actions (Fig.~\ref{fig:dataclaw}).

\vspace{-4pt}
\subsection{Processing Pipeline and Action Representation}
\label{sec:processing}
We convert heterogeneous data sources into a common wrist-camera observation--action schema by synchronizing video, camera poses, and gripper states at 30\,Hz. We retain available calibration for image rectification and geometry-aware modeling, reject invalid samples, and divide trajectories into aligned training windows. Appendix~\ref{app:processing} details ingestion, resampling, rectification, and validation.

To remove dependence on global coordinates and support open-world recordings whose wrists are tracked in separate local frames, we express each trajectory relative to the first frame $t_0$ of its window. For arm $i\in\{L,R\}$ with camera pose $(\mathbf R_t^i,\mathbf p_t^i)$ in its available reference frame, we jointly define the relative pose and 10D action as:
\begin{equation}
\begin{gathered}
\Delta\mathbf p_t^i=(\mathbf R_{t_0}^i)^\top(\mathbf p_t^i-\mathbf p_{t_0}^i),
\qquad
\Delta\mathbf R_t^i=(\mathbf R_{t_0}^i)^\top\mathbf R_t^i,\\
\mathbf a_t^i=\bigl[(\Delta\mathbf p_t^i)^\top,\,
\operatorname{rot6d}(\Delta\mathbf R_t^i)^\top,\,g_t^i\bigr]^\top\in\mathbb R^{10},
\qquad
\operatorname{rot6d}(\mathbf R)=\bigl[\mathbf R_{:,1}^\top,\,\mathbf R_{:,2}^\top\bigr]^\top.
\end{gathered}
\label{eq:per-arm-action}
\end{equation}
We use the continuous 6D rotation representation of \citet{zhou2019continuity}. For tweezer-style devices, $g_t^i$ is the recorded opening angle, termed gripper width and used without temporal differencing.

\vspace{-6pt}
\section{UMI-WM: Interactive Multi-View World Modeling}
\label{sec:method-overview}

\vspace{-4pt}
\subsection{Action Conditioning}
\label{sec:wm-action}
Let $\mathcal O_{1:H}$ be the paired-wrist observation history, $\mathcal A=\{\mathbf a_{1:T}^{L},\mathbf a_{1:T}^{R}\}$ the per-arm actions (Eq.~\ref{eq:per-arm-action}), and $\mathcal G$ the available camera intrinsics and extrinsics. We initialize a video DiT from Matrix-Game 3.5~\citep{qian2026matrixgame35}. A frozen VAE encodes each view; a shared temporal transformer encodes the aligned per-arm actions. Conditional prediction and view-specific action injection are:
\begin{equation}
p_\theta\!\left(\mathcal O_{H+1:T}^{L,R}\mid \mathcal O_{1:H},\mathcal A,\mathcal G\right),\qquad \widetilde{\mathbf h}_{v,n,s}^{(\ell)}=\mathbf h_{v,n,s}^{(\ell)}+\gamma_\ell\mathbf u_n^{\pi(v)},
\label{eq:wm-action-conditioning}
\end{equation}
where $\mathcal O_{H+1:T}^{L,R}$ denotes the paired future wrist videos; $\mathbf h_{v,n,s}^{(\ell)}$ is the visual token at view $v$, latent time index $n$, spatial position $s$, and layer $\ell$; $\pi(v)$ maps the view to its associated arm; $\mathbf u_n^{\pi(v)}$ is the encoded action state; and $\gamma_\ell$ is a learned scalar gate. The action state is broadcast over that view's spatial tokens before each DiT block, while the hub exchanges information across the two arms.

\begin{figure}[!htbp]
\centering
\setlength{\abovecaptionskip}{4pt}
\includegraphics[width=\linewidth]{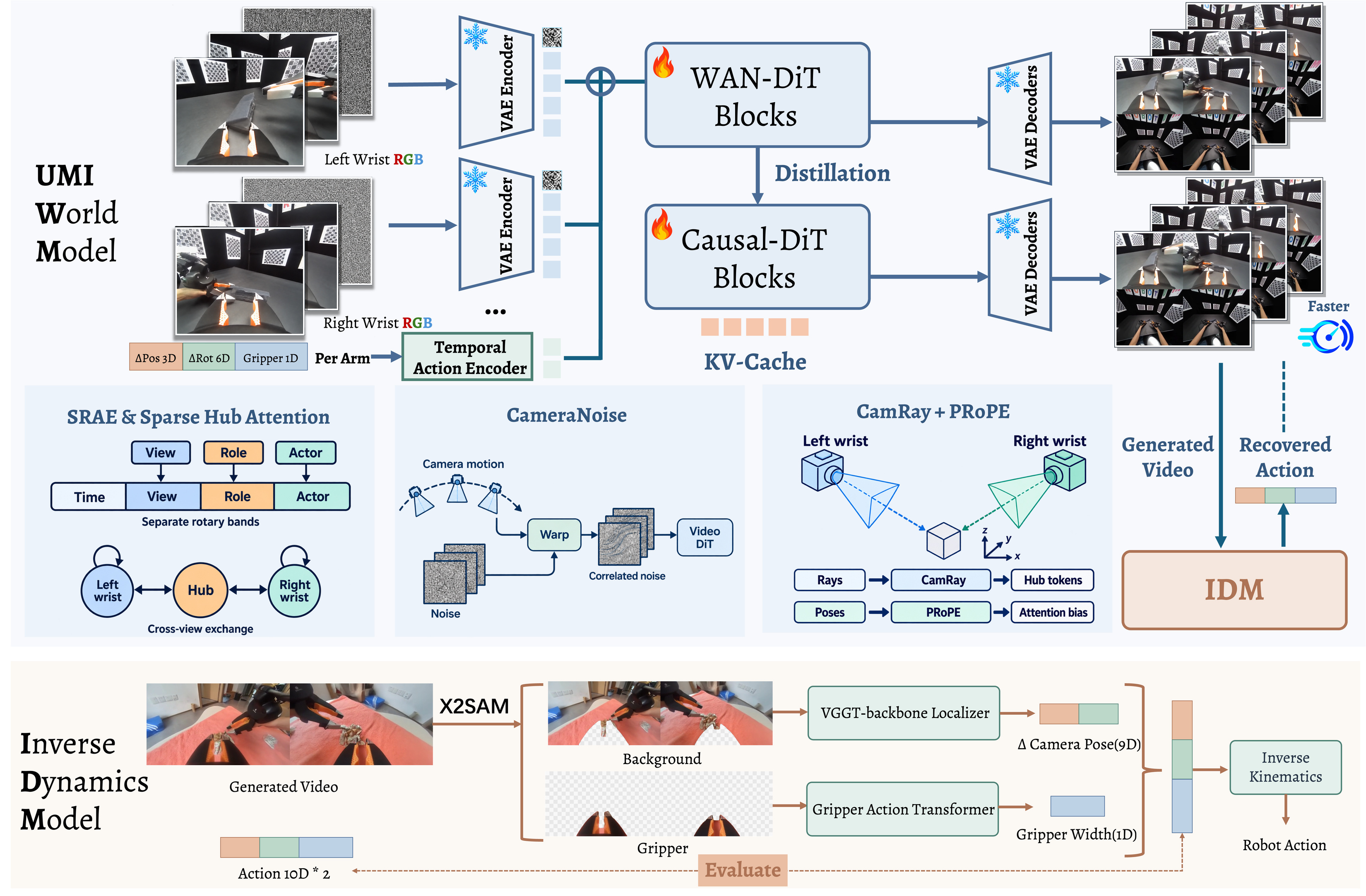}
\caption{\textbf{UMI-WM overview.} SRAE and Sparse Hub Attention coordinate wrist streams; CameraNoise supplies motion-aware noise; CamRay and PRoPE provide geometric conditioning. Distillation enables four-step causal generation. The IDM recovers actions from generated videos for BAF evaluation and robot execution via inverse kinematics.}
\label{fig:method-overview}
\vspace{-4pt}
\end{figure}

\vspace{-4pt}
\subsection{Multi-View Modeling and Causal Generation}
\label{sec:wm-cross-view}
\textbf{SRAE and Sparse Hub Attention.} Adapting Simplex Rotary Agent Encoding (SRAE) and Sparse Hub Attention from Gamma-World~\citep{liu2026gammaworld}, we encode view, role, and actor identities separately and exchange information between wrist streams through learned hub tokens. Actor identity routes the appropriate per-arm action state in Eq.~\ref{eq:wm-action-conditioning}, while the hub receives a validity-masked summary of both arms, preserving view-specific control and shared-scene context. Appendix~\ref{app:encoding} details tokenization, rotary identity construction, attention connectivity, and action routing.

\vspace{-3pt}
\textbf{Camera-aware geometry conditioning.} We combine CameraNoise~\citep{zhao2026cameranoise} with ray-based and relative-pose conditioning inspired by PRoPE~\citep{li2025cameras} and Matrix-Game 3.5~\citep{qian2026matrixgame35}, coupling per-view motion control with cross-view geometric exchange. This interface supports both local-frame and calibrated shared-frame trajectories; Appendix~\ref{app:camera} details the noise construction, geometric features, and attention biases.

\vspace{-3pt}
\textbf{Progressive training and causal generation.} Training follows the data pyramid from broad single-wrist coverage to joint dual-wrist modeling and calibrated geometry refinement. Following CausVid~\citep{yin2025causvid} and Self Forcing~\citep{huang2025selfforcing}, we then convert the model to block-causal generation and distill it to four denoising steps for interactive rollout. Appendix~\ref{app:objectives} details the training stages, objectives, causal masks, distillation, and inference.

\vspace{-4pt}
\subsection{Inverse Dynamics Model and Bimanual Action Fidelity}
\label{sec:idm}
\textbf{Inverse dynamics model.} Similar to VPT~\citep{baker2022vpt}, our IDM recovers per-arm relative camera poses and gripper widths from generated wrist videos (Fig.~\ref{fig:method-overview}). X2SAM~\citep{wang2026x2sam} masks separate the gripper foreground from the scene background. For pose recovery, a trainable temporal head processes frozen VGGT~\citep{wang2025vggt} background features. For gripper estimation, a temporal transformer processes fixed RGB crops that preserve local appearance, using the foreground masks as auxiliary inputs. We train both branches on real UMI videos from the training split only, with supervision from recorded translations, rotations, gripper openings, and temporal changes in opening; the IDM training set shares no episodes with the 320 held-out test episodes used in Section~\ref{sec:experiments}. At each conditioning timestamp, the predicted pose relative to the initial frame and the gripper width form the 10D action in Eq.~\ref{eq:per-arm-action}.

\textbf{Bimanual action fidelity.} For arm $i\in\{L,R\}$ and component $k\in\{\mathrm{pos},\mathrm{rot},\mathrm{grip}\}$, let $E_k^i$ be the mean error between conditioning and video-recovered actions. We define:
\begin{equation}
\mathrm{BAF}=100\exp\!\left[-\frac{\lambda}{6}\sum_{i\in\{L,R\}}\sum_k\left[\frac{E_k^i}{\tau_k}-1\right]_{+}\right],
\label{eq:idm-action-consistency}
\end{equation}
where $[x]_{+}=\max(x,0)$. Higher scores indicate closer action agreement, reaching 100 when all six mean errors lie within their tolerance bands. We use $\tau=(5\,\mathrm{cm},5^\circ,0.1\,\mathrm{rad})$ and $\lambda=0.25$ for all models; Appendix~\ref{app:idm} details component errors and score aggregation over valid predictions.

\vspace{-6pt}
\section{Experiments}
\label{sec:experiments}
We compare UMI-WM with baselines (Table~\ref{tab:main-results}), probe its components (Table~\ref{tab:ablations}), and evaluate Bimanual Action Fidelity (BAF) on GT and generated wrist videos (Table~\ref{tab:idm-recovery}).

\vspace{-4pt}
\subsection{Experimental Setup}
\label{sec:exp-setup}
\textbf{Test data.} We evaluate UMI-WM on 320 UMI test episodes comprising 1,008 paired-wrist windows, each containing 189 frames with per-arm controls at 30\,Hz. Task descriptions are taken from existing annotations or generated from wrist-camera videos before evaluation. Scores are averaged over two generation seeds with equal episode weights.

\vspace{-3pt}
\textbf{Evaluation metrics.} We evaluate visual fidelity using PSNR, SSIM~\citep{wang2004ssim}, LPIPS~\citep{zhang2018perceptual}, FID~\citep{heusel2017fid}, and FVD~\citep{unterthiner2018fvd}, video quality using VBench~\citep{huang2024vbench} and WorldArena~\citep{shang2026worldarenaunifiedbenchmarkevaluating} (Appendix~\ref{app:baseline-comparisons}), and action following using BAF (Appendix~\ref{app:baf-analysis}). We compare generated frames with the corresponding ground-truth frames, excluding the initial conditioning frame; FID and FVD use one selected window per episode. We report paired-wrist FPS for generation speed.

\begin{table}[H]
\centering
\caption{\textbf{Comparison with baseline world models.} Bold/underline: best/second-best. GT is unranked; -- means not reported. Gamma-World is fine-tuned on our UMI training split. VBench: SC/BC subject/background consistency, MS motion smoothness, IQ imaging quality. WorldArena: JS JEPA similarity, PS photometric smoothness, PV perspectivity. FPS: paired frames/s.}
\label{tab:main-results}
\begingroup
\fontsize{8.5}{10}\selectfont
\setlength{\tabcolsep}{2.2pt}
\renewcommand{\arraystretch}{1.1}
\resizebox{\linewidth}{!}{%
\begin{tabular}{@{}l*{13}{c}@{}}
\toprule
& \multicolumn{5}{c}{\cellcolor{benchgreen}\textbf{Reference and distribution fidelity}} & \multicolumn{4}{c}{\cellcolor{benchpeach}\textbf{VBench}} & \multicolumn{3}{c}{\cellcolor{benchblue}\textbf{WorldArena}} & \multicolumn{1}{c@{}}{\cellcolor{benchgray}\textbf{Eff.}} \\
\cmidrule(lr){2-6}\cmidrule(lr){7-10}\cmidrule(lr){11-13}\cmidrule(l){14-14}
Model & PSNR$\uparrow$ & SSIM$\uparrow$ & LPIPS$\downarrow$ & FID$\downarrow$ & FVD$\downarrow$ & SC$\uparrow$ & BC$\uparrow$ & MS$\uparrow$ & IQ$\uparrow$ & JS$\uparrow$ & PS$\uparrow$ & PV$\uparrow$ & FPS$\uparrow$ \\
\midrule
GT video & -- & -- & -- & -- & -- & 0.860 & 0.926 & 0.984 & 0.716 & 1.000 & 3.377 & 3.750 & -- \\
\addlinespace[2pt]
\multicolumn{14}{@{}l}{\textit{Text-conditioned world models}} \\
Wan2.2 TI2V-5B & 8.345 & 0.312 & 0.715 & 99.76 & 3477.06 & 0.824 & 0.903 & \textbf{0.988} & 0.604 & 0.059 & 1.827 & \textbf{3.875} & 0.18 \\
Cosmos Predict2.5 & 7.759 & 0.308 & 0.685 & 98.62 & 967.40 & 0.688 & 0.850 & 0.974 & 0.631 & 0.284 & 1.747 & \underline{3.500} & 0.10 \\
MiniMax H3 & 9.477 & 0.371 & 0.577 & 58.91 & 687.94 & 0.768 & 0.874 & 0.982 & 0.609 & 0.080 & 1.772 & 3.375 & 0.70 \\
\addlinespace[2pt]
\multicolumn{14}{@{}l}{\textit{Action-conditioned world model}} \\
Gamma-World & 12.264 & 0.548 & 0.397 & 40.30 & 331.25 & 0.874 & \underline{0.933} & \underline{0.984} & 0.655 & 0.382 & 4.236 & 3.375 & 0.55 \\
\addlinespace[2pt]
\multicolumn{14}{@{}l}{\textit{Camera-controlled world models}} \\
HY-WorldPlay 5B & 8.599 & 0.338 & 0.606 & 68.15 & 638.08 & 0.825 & 0.899 & 0.968 & \underline{0.705} & 0.135 & 2.296 & 3.250 & 0.81 \\
LingBot-World-Fast & 7.660 & 0.270 & 0.752 & 128.16 & 2190.36 & 0.627 & 0.855 & 0.958 & 0.464 & 0.179 & 1.669 & 3.250 & 2.32 \\
\midrule
\multicolumn{14}{@{}l}{\textit{UMI-WM (ours)}} \\
\textbf{UMI-WM Teacher} & \textbf{12.513} & \underline{0.576} & \textbf{0.332} & \textbf{33.52} & \textbf{232.53} & \textbf{0.885} & \textbf{0.938} & \underline{0.984} & \textbf{0.707} & \underline{0.386} & \textbf{4.339} & \underline{3.500} & 0.26 \\
\textbf{UMI-WM Student} & \underline{12.334} & \textbf{0.580} & \underline{0.336} & \underline{37.89} & \underline{284.68} & \underline{0.875} & \underline{0.933} & \underline{0.984} & 0.655 & \textbf{0.408} & \underline{4.300} & 3.125 & 1.81 \\
\bottomrule
\end{tabular}%
}
\endgroup
\end{table}

\vspace{-4pt}
\subsection{Baseline Comparisons}
\label{sec:exp-generation}
\looseness=-1
Table~\ref{tab:main-results} compares UMI-WM with text-, action-, and camera-conditioned baselines. Text-conditioned models such as MiniMax-H3~\citep{minimax2026h3} receive task descriptions and initial wrist images; camera-controlled models additionally receive mapped wrist trajectories. Both groups generate wrists separately. Gamma-World is fine-tuned on our UMI training split and uses paired 20D actions encoding each arm's relative camera pose and gripper width (Eq.~\ref{eq:per-arm-action}).

\begin{figure}[!htb]
\centering
\setlength{\abovecaptionskip}{4pt}
\begin{tikzpicture}
\node[inner sep=0, anchor=south west] (img) {\includegraphics[width=\linewidth]{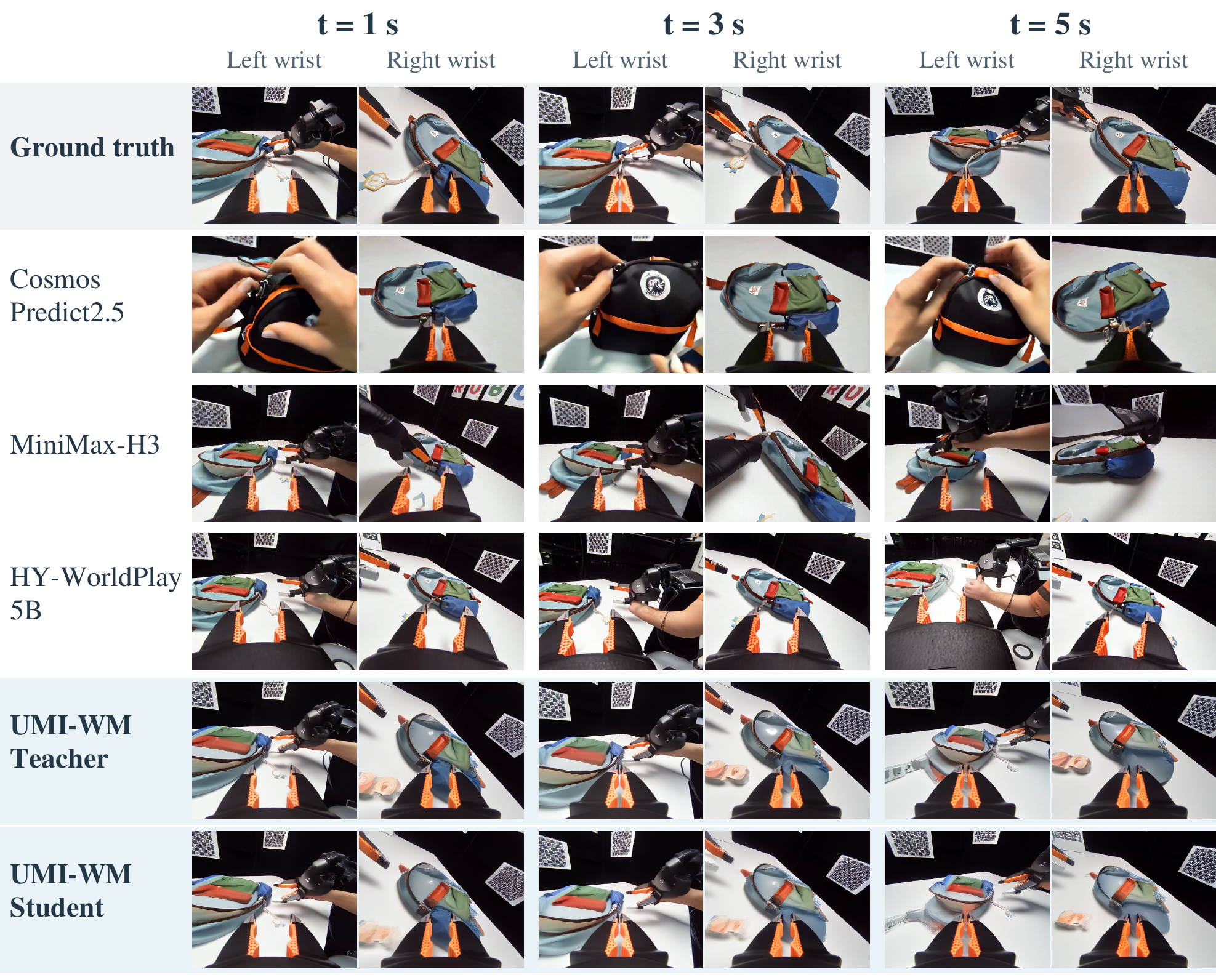}};
\begin{scope}[x={(img.south east)}, y={(img.north west)}]
\tikzset{fail/.style={draw=red, line width=0.9pt}, flabel/.style={fill=white, fill opacity=0.85, text opacity=1, text=red, inner sep=1pt, font=\fontsize{5.5}{6}\selectfont\bfseries, align=center}}
\draw[fail] (0.158,0.620) rectangle (0.294,0.759);
\node[flabel, anchor=south] at (0.226,0.623) {hands replace\\gripper};
\draw[fail] (0.865,0.468) rectangle (0.9985,0.607);
\node[flabel, anchor=south] at (0.932,0.471) {own gripper\\vanishes};
\draw[fail] (0.295,0.316) rectangle (0.431,0.456);
\node[flabel, anchor=south] at (0.363,0.319) {gripper open\\(GT: closed)};
\end{scope}
\end{tikzpicture}
\caption{\textbf{Paired-wrist qualitative comparison.} Rows show GT and models; columns show three timestamps, each pairing left and right wrist views. Red boxes mark baseline failures.}
\label{fig:qualitative-comparison}
\vspace{-7pt}
\end{figure}

\vspace{-3pt}
Figure~\ref{fig:qualitative-comparison} compares UMI-WM with Cosmos Predict2.5, MiniMax-H3, and HY-WorldPlay 5B at 1, 3, and 5\,s. Teacher and Student maintain the backpack appearance and paired-wrist layout more consistently across the sequence. Further qualitative comparisons appear in Appendix~\ref{app:baseline-comparisons}.

\vspace{-3pt}
\textbf{Generation quality.} The Teacher achieves the best PSNR, LPIPS, FID, and FVD, and the Student the best SSIM. Gamma-World, the strongest baseline, reaches an FVD of 331.25 and an LPIPS of 0.397. Across the reported VBench and WorldArena metrics, at least one UMI-WM variant ranks first or ties for second. The gains in frame reconstruction and distributional fidelity demonstrate UMI-WM's advantage in wrist-view prediction over the compared baselines.

\vspace{-3pt}
\textbf{Excluded benchmark metrics.} We also evaluated WorldModelBench~\citep{Li2025WorldModelBench}, iWorld-Bench~\citep{fang2026iworldbench}, and EWMBench~\citep{hu2025ewmbench}, but omit them from the main comparison because their tested metrics were unreliable on our wrist videos: static or noisy controls could receive favorable scores, pose estimates were unstable, and the released gripper tracker often failed. Appendix~\ref{app:baseline-comparisons} summarizes these checks and the resulting metric selection.

\vspace{-3pt}
\textbf{Generation efficiency.} Timings use one A100-80GB GPU and include conditioning, denoising, and decoding for both wrists. The four-step Student reaches 1.81 paired frames/s versus 0.26 for the 30-step Teacher, a $6.9\times$ speedup, for joint generation of 189-frame wrist videos. LingBot-World-Fast reaches 2.32 paired frames/s by generating each wrist separately. The Student combines strong visual prediction quality with substantially faster paired-wrist generation than the Teacher.

\vspace{-4pt}
\subsection{Geometry Ablations and Sampling Sensitivity}
\label{sec:exp-ablations}
\looseness=-1
\textbf{Geometry probes.} Removing CamRay with geometric hub routing or PRoPE lowers BAF from 76.36 to 46.79 or 47.26, respectively, while also worsening LPIPS and imaging quality. Both variants start from the Stage~2 checkpoint at step 15,000 and train the retained adapters, with the backbone and action conditioner frozen. Within these probes, the full geometry configuration achieves the best visual and action fidelity, supporting our design for geometry conditioning.

\textbf{Student sampling.} With the checkpoint, observations, actions, seeds, and horizons fixed, four steps achieve the best LPIPS of 0.336 with BAF 71.55 at 1.81 paired frames/s. Two steps increase FPS to 3.31 but reduce BAF to 53.82; eight steps raise BAF to 76.12 and improve imaging quality but halve FPS to 0.94. This supports four steps as our default balance between action fidelity and speed.

\begin{table}[!htb]
\centering
\setlength{\abovecaptionskip}{4pt}
\caption{\textbf{Geometry probes and sampling sensitivity.} Bold/underline: best/second-best values within the Student sampling block for each metric.}
\label{tab:ablations}
\begingroup
\footnotesize
\setlength{\tabcolsep}{3pt}
\renewcommand{\arraystretch}{1.0}
\begin{tabular*}{\linewidth}{@{\extracolsep{\fill}}lcccc@{}}
\toprule
Variant & LPIPS $\downarrow$ & \shortstack{Imaging\\quality $\uparrow$}
& BAF $\uparrow$ & \shortstack{Paired-wrist\\FPS $\uparrow$} \\
\midrule
UMI-WM Teacher & 0.332 & 0.707 & 76.36 & 0.26 \\
\addlinespace[3pt]
\rowcolor{benchblue}
\multicolumn{5}{l}{\textbf{Geometry probes}} \\
w/o CamRay + geometric Hub & 0.456 & 0.654 & 46.79 & 0.65 \\
w/o PRoPE & 0.434 & 0.655 & 47.26 & 0.81 \\
\rowcolor{benchpeach}
\multicolumn{5}{l}{\textbf{Student sampling}} \\
Student, 2 steps & 0.416 & 0.638 & 53.82 & \textbf{3.31} \\
\textbf{Student, 4 steps (default)} & \textbf{0.336} & \underline{0.655} & \underline{71.55} & \underline{1.81} \\
Student, 8 steps & \underline{0.385} & \textbf{0.673} & \textbf{76.12} & 0.94 \\
\bottomrule
\end{tabular*}
\endgroup
\vspace{-8pt}
\end{table}

\subsection{Bimanual Action Fidelity Analysis}
\vspace{-8pt}
\label{sec:exp-idm}
Table~\ref{tab:idm-recovery} reports per-wrist motion errors (Appendix~\ref{app:idm}) and BAF (Eq.~\ref{eq:idm-action-consistency}). All models share the same frozen IDM and query timestamps. We compute BAF per window and IDM seed, then average seeds within windows and windows within episodes, weighting episodes equally. The Teacher and Student reach BAF 76.36 and 71.55, above adapted Gamma-World at 52.93, the best camera-controlled baseline at 41.26, and the static reference at 18.78. UMI-WM leads most error columns; Matrix-Game 3.0 has lower right-wrist position error (15.59\,cm versus the Teacher's 15.63\,cm). Text-conditioned scores remain near the static reference. These results support UMI-WM's advantage in generating wrist-view motion that agrees with the reference actions for both arms.

BAF adds a monotonic action-fidelity scale to the visual and distributional metrics above, anchored by GT at 97.72 and static at 18.78: higher scores closer to GT indicate stronger agreement with conditioning actions, whereas lower scores closer to static indicate weaker controllable motion. Scores below static may indicate a collapsed video with little effective information; for camera- and text-conditioned baselines, BAF measures agreement with recorded actions (Appendix~\ref{app:baf-analysis}).

\begin{table}[H]
\centering
\setlength{\abovecaptionskip}{4pt}
\caption{\textbf{Bimanual Action Fidelity comparison.} Bold/underline: best/second-best among generated models; GT and static controls are references. Gamma-World is fine-tuned on our UMI training split; all entries use the shared 320-episode test set and the same frozen IDM.}
\label{tab:idm-recovery}
\begingroup
\setlength{\aboverulesep}{1.5pt}
\setlength{\belowrulesep}{1.5pt}
\footnotesize
\setlength{\tabcolsep}{1.7pt}
\renewcommand{\arraystretch}{1.0}
\begin{tabular*}{\linewidth}{@{\extracolsep{\fill}}l*{7}{c}@{}}
\toprule
& \multicolumn{3}{c}{Left wrist ($L$)}
& \multicolumn{3}{c}{Right wrist ($R$)} & \\
\cmidrule(lr){2-4}\cmidrule(lr){5-7}
Video source / model
& \shortstack{$E_{\mathrm{pos}}^L\!\downarrow$\\(cm)}
& \shortstack{$E_{\mathrm{rot}}^L\!\downarrow$\\(${}^\circ$)}
& \shortstack{$E_{\mathrm{grip}}^L\!\downarrow$\\(rad)}
& \shortstack{$E_{\mathrm{pos}}^R\!\downarrow$\\(cm)}
& \shortstack{$E_{\mathrm{rot}}^R\!\downarrow$\\(${}^\circ$)}
& \shortstack{$E_{\mathrm{grip}}^R\!\downarrow$\\(rad)}
& \shortstack{BAF\\$\uparrow$} \\
\midrule
GT video (shared test) & 3.89 & 3.45 & 0.0275 & 4.52 & 4.10 & 0.0231 & 97.72 \\
Static conditioning frame (control) & 23.01 & 77.21 & 0.0762 & 34.43 & 85.07 & 0.0915 & 18.78 \\
\midrule
\multicolumn{8}{l}{\textit{Text-Conditioned World Models}} \\
Wan2.2 TI2V-5B & 22.69 & 77.28 & 0.1011 & 34.46 & 85.02 & 0.1168 & 18.64 \\
Cosmos Predict2.5-2B & 32.75 & 85.62 & 0.1571 & 35.45 & 88.04 & 0.1625 & 17.54 \\
MiniMax-H3 & 21.19 & 83.12 & 0.1059 & 39.02 & 101.95 & 0.1085 & 15.20 \\
Cosmos3-Nano-Policy-DROID (I2V) & 28.46 & 106.11 & 0.1250 & 37.84 & 75.43 & 0.1568 & 14.78 \\
RynnWorld-Teleop (text+image SFT) & 22.93 & 77.81 & 0.1037 & 33.59 & 84.26 & 0.1273 & 18.94 \\
\midrule
\multicolumn{8}{l}{\textit{Action-Conditioned World Models}} \\
Genie-Envisioner (GE-Sim) & 21.63 & 72.97 & 0.0855 & 35.84 & 79.42 & 0.0621 & 20.57 \\
RynnWorld-Teleop (control band) & 22.41 & 77.50 & 0.1171 & 33.88 & 85.14 & 0.1224 & 18.72 \\
Gamma-World & 19.42 & 25.27 & 0.0808 & 24.39 & 28.70 & 0.0771 & 52.93 \\
\midrule
\multicolumn{8}{l}{\textit{Camera-Controlled World Models}} \\
HY-WorldPlay 5B & 16.66 & 42.11 & 0.1264 & 25.33 & 59.51 & 0.1726 & 36.53 \\
LingBot-World-Fast & 38.71 & 56.68 & 0.1411 & 30.74 & 65.96 & 0.1444 & 24.19 \\
DreamX-World-5B-Cam & 49.53 & 102.66 & 0.1249 & 28.56 & 70.63 & 0.1208 & 14.89 \\
Matrix-Game 2.0 & 22.72 & 51.33 & 0.1349 & 23.78 & 57.52 & 0.1070 & 32.36 \\
Matrix-Game 3.0 & 20.33 & 37.53 & 0.1424 & \textbf{15.59} & 55.24 & 0.1712 & 41.26 \\
YUME 1.5 & 23.18 & 67.70 & 0.1048 & 31.83 & 74.95 & 0.1511 & 22.51 \\
Infinite-World & 19.88 & 63.08 & 0.1345 & 31.62 & 66.66 & 0.1685 & 25.23 \\
\midrule
\textbf{UMI-WM Teacher} & \textbf{12.49} & \underline{6.66} & \underline{0.0285} & \underline{15.63} & \textbf{18.88} & \underline{0.0188} & \textbf{76.36} \\
\textbf{UMI-WM Student} & \underline{14.44} & \textbf{6.16} & \textbf{0.0270} & 17.91 & \underline{24.78} & \textbf{0.0165} & \underline{71.55} \\
\bottomrule
\end{tabular*}
\endgroup

\end{table}

We map each baseline's available controls to the common per-arm action space before applying the same frozen IDM and BAF computation. In Table~\ref{tab:idm-recovery}, ``I2V'' means image-to-video input, ``text+image SFT'' is RynnWorld-Teleop's supervised text-and-image route, and ``GE-Sim'' is Genie-Envisioner's rendered-control interface. RynnWorld-Teleop's ``control band'' renders mapped controls as a dual-hand skeleton video (Appendix~\ref{app:baf-analysis}).

\vspace{-4pt}
\subsection{Applications}
\label{sec:exp-realtime}
\begin{figure}[!htb]
\centering
\setlength{\abovecaptionskip}{4pt}
\includegraphics[width=\linewidth]{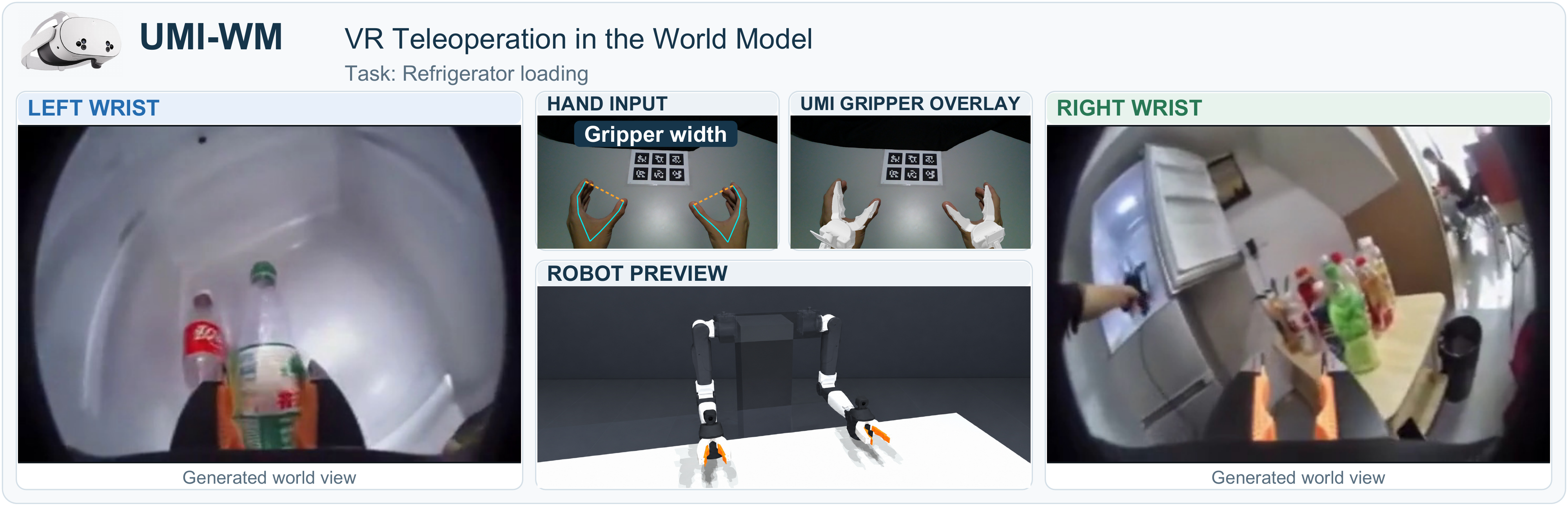}
\caption{\textbf{VR teleoperation in the world model.} Interface for a refrigerator-loading rollout with hand tracking, a UMI gripper overlay, generated wrist views, and a robot preview.}
\label{fig:teleop-workflow}
\vspace{4pt}
\end{figure}

\textbf{VR teleoperation in the world model.} Figure~\ref{fig:teleop-workflow} demonstrates UMI-WM as a visual environment for bimanual teleoperation. The interface maps tracked hand motion and finger separation to per-arm camera-pose and gripper controls, then jointly generates both wrist views as visual feedback for subsequent commands. Hand tracking and a UMI gripper overlay visualize the input mapping, while the IDM recovers generated actions for inverse-kinematics mapping to robot configurations.

\textbf{Robot action planning.} To validate the interface, we execute a UMI-WM rollout on the physical robot. After interactive approval, the IDM recovers relative wrist-camera poses and gripper widths. A calibrated mapping converts the poses into end-effector targets for inverse kinematics, while the recovered widths drive the grippers. The matched physical and wrist-view sequences from initial approach to completed transport are shown in Appendix~\ref{app:applications} (Fig.~\ref{fig:robot-execution}).

\vspace{-3pt}
\textbf{Visual domain randomization.} Similar to ReShoot~\citep{kim2026reshootgenerativevisualdomain}, UMI-WM regenerates paired wrist videos after consistent appearance edits to both initial frames, keeping per-arm trajectories and gripper controls fixed. The edits expand the visual coverage of the demonstrations, and BAF decides which edited rollouts join policy training alongside the originals. Fig.~\ref{fig:app-domain-rand} in Appendix~\ref{app:applications} shows hue, object-color, and background edits to both initial frames, the rollout regenerated from each, and its BAF relative to the unedited rollout.

\vspace{-3pt}
\textbf{Data quality screening.} UMI-WM also audits recorded data: the frozen IDM recovers actions from raw wrist-camera frames, and segments where they persistently depart from the recorded SLAM trajectories and gripper controls are flagged for manual review, trajectory re-estimation, or temporal realignment. Fig.~\ref{fig:app-screening} in Appendix~\ref{app:applications} shows per-timestep discrepancy traces for a clean recording and two corrupted controls, screening scores over the test windows, and a flagged real recording whose recorded and recovered wrist translations and 3D paths diverge.

\vspace{-6pt}
\section{Conclusion}
\label{sec:conclusion}
We introduced UMI-WM, an interactive world model that jointly predicts wrist-camera videos from per-arm camera motion and gripper controls. A unified UMI data pipeline, cross-view information exchange, and camera-aware conditioning support few-step causal generation. On the shared 320-episode evaluation, the Teacher leads PSNR, LPIPS, FID, and FVD, while the Student leads SSIM and reaches 1.81 paired-wrist FPS, a $6.9\times$ speedup. Their BAF scores of 76.36 and 71.55 lie between the GT reference at 97.72 and the static reference at 18.78, and exceed all compared baselines on the higher-is-better action-fidelity scale. Removing camera-aware conditioning in the geometry probes lowers both imaging quality and BAF, supporting its role in action following. Together with the physical execution demonstration, these results connect wrist-view prediction, action recovery, and robot planning through a shared control interface.

\clearpage
\section*{AI Use Statement}
Generative AI tools were used to aid and polish writing, support literature retrieval and discovery, and assist research ideation and execution, including experiment planning, code development and debugging, data-processing utilities, and figure preparation. The authors reviewed suggestions against cited sources, implementation code, experimental records, and visual results and take responsibility for the final manuscript and its claims.

\section*{Ethics Statement}
Our UMI data were collected internally for research on visual world modeling. Wrist-camera and egocentric recordings underwent privacy screening before use, and the paper reports aggregate evaluations together with selected task-scene visualizations. Public datasets and third-party assets are used only under their respective licenses and permissions.

\section*{Reproducibility Statement}
Sections~\ref{sec:data}--\ref{sec:experiments} describe the data sources and processing, coordinate and action conventions, model components, staged training, IDM and BAF, and evaluation protocols. The appendices expand the model objectives, geometry-probe and Gamma-World training settings, benchmark configurations, metric aggregation, validity handling, and evaluator limitations.

\section*{Acknowledgements}
This work was supported by the NUS Start-up Grant (A-0010106-00-00), Razer--NUS Joint Lab, Romoya--NUS Joint Initiative, and NAII Seed Funding (A-8004365-00-00).

\bibliography{references}
\bibliographystyle{iclr2027_conference}

\clearpage
\appendix
\section{Data Collection and Processing}
\label{app:data}

\subsection{UMI Data Collection}
\label{app:umi-collection}
\begin{figure}[htbp]
\centering
\includegraphics[width=\linewidth]{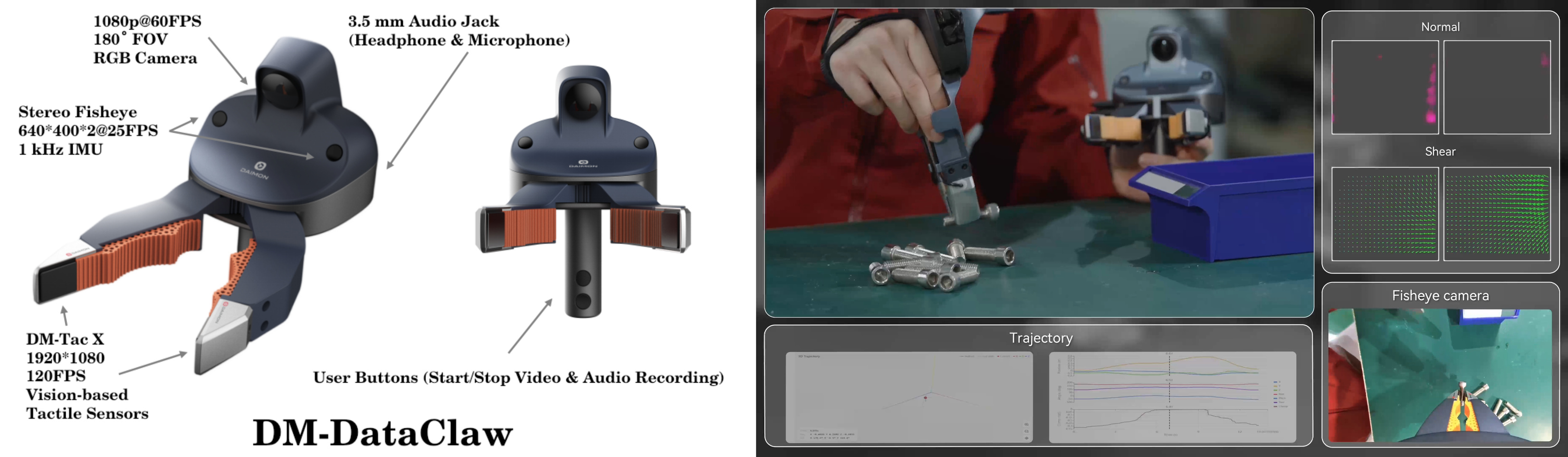}
\caption{\textbf{DM-DataClaw hardware and data collection.} Device specifications and a handheld collection example, adapted from Daimon-Infinity~\citep{daimon2026infinity}.}
\label{fig:dataclaw-hardware}
\end{figure}

DM-DataClaw is the tweezer-style UMI gripper used in our collection: its measured jaw opening supplies the gripper-width component of Eq.~\ref{eq:per-arm-action}. The camera and jaws are rigidly coupled, so wrist-view motion and jaw state share a device timeline. Figure~\ref{fig:dataclaw-hardware} shows the camera and tactile hardware together with a representative handheld session. Tactile and force sensing provide complementary information for contact-rich manipulation~\citep{shan2026learningphysicalinteractionsurvey}.

\begin{figure}[htbp]
\centering
\includegraphics[width=0.60\linewidth,trim=0bp 250bp 0bp 300bp,clip]{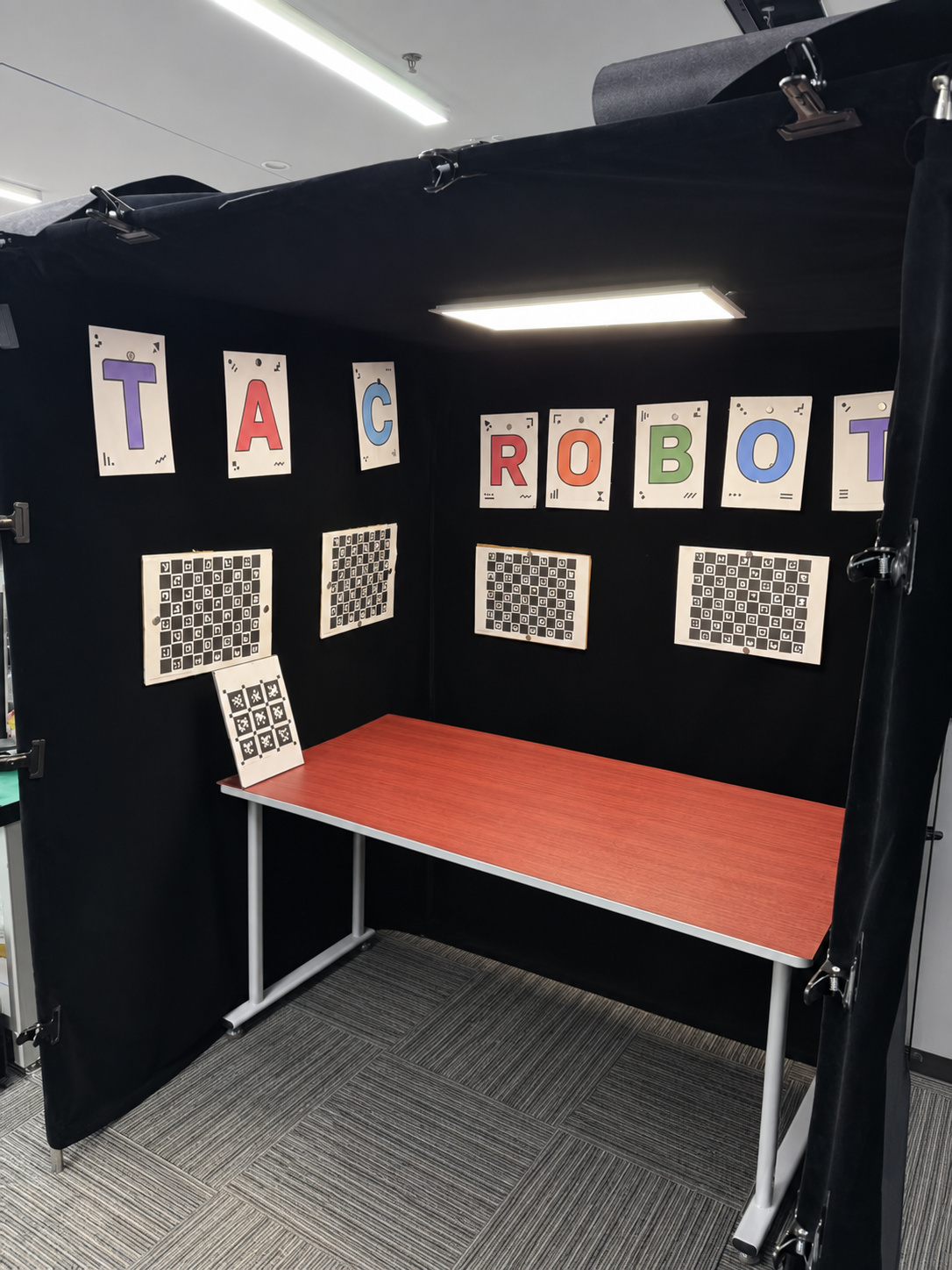}
\caption{\textbf{UMI data collection booth.} Studio enclosure and calibration targets.}
\label{fig:umi-collection-booth}
\end{figure}

\textbf{Collection environment.} Figure~\ref{fig:umi-collection-booth} shows the studio enclosure. AprilGrid targets~\citep{furgale2013kalibr}, built from AprilTag markers~\citep{olson2011tags}, span multiple viewing directions to preserve geometric constraints as the operator moves within the booth.

\begin{figure}[H]
\centering
\includegraphics[width=\linewidth]{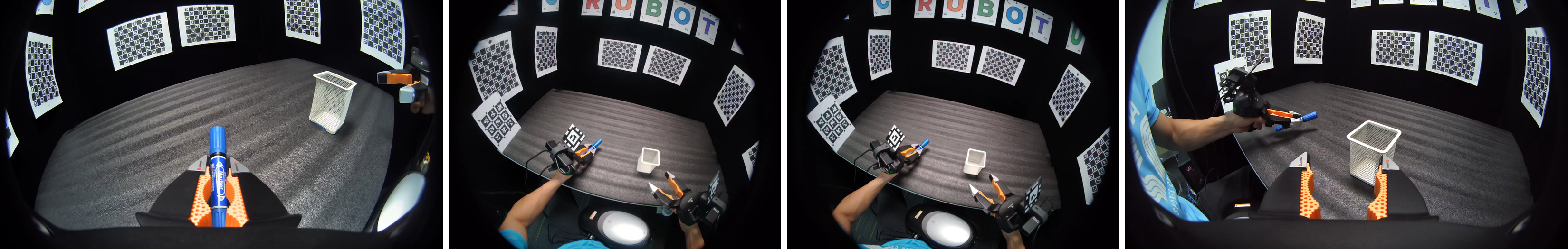}
\caption{\textbf{Four-camera studio recording.} Left wrist, left eye, right eye, and right wrist.}
\label{fig:studio-four-camera}
\end{figure}

\textbf{Stereo egocentric views.} The synchronized eye-camera pair provides wider and more persistent AprilGrid visibility for multi-frame registration of all camera trajectories in a shared world frame (Fig.~\ref{fig:studio-four-camera}); fixed extrinsics then recover both eye-camera poses.

\subsection{Processing Pipeline Details}
\label{app:processing}
\textbf{Source-specific ingestion.} We convert heterogeneous recordings into a common episode format containing wrist-camera videos, pose trajectories, and gripper states, together with available egocentric observations and calibration metadata. Source-specific adapters standardize field conventions while preserving source identifiers, episode indices, and video-segment mappings, allowing each processed sample to be traced back to its original recording.

\textbf{Temporal alignment.} We align video and state streams using their timestamps and resample them onto a common 30\,Hz timeline. Translations are linearly interpolated, rotations use spherical linear interpolation (SLERP; \citealp{shoemake1985animating}), and gripper readings are aligned through zero-order hold. Video streams are resampled by dropping or repeating frames to match the aligned state sequences.

\textbf{Action serialization.} We apply Eq.~\ref{eq:per-arm-action} after temporal resampling, assigning each retained video frame one pose--gripper tuple. Source and segment mappings remain attached to the converted sequence so it can be checked against the original streams before training windows are emitted.

\textbf{Image rectification and calibration.} We rectify fisheye wrist images to pinhole views at training resolution and retain their intrinsics, hand--eye extrinsics, and available shared-frame registration.

\textbf{Quality checks.} We validate temporal coverage and frame counts, video decodability, pose and action validity, and camera calibration. Checks reject missing required values, non-finite states, invalid rotations, and inconsistent episode indexing. Geometry validation additionally checks camera intrinsics, rigid transforms, and black borders introduced by rectification.

\section{Model and Training Details}
\label{app:training}

\subsection{Action, Identity, and Hub Encoding}
\label{app:encoding}
\textbf{Video and action tokenization.} A frozen video VAE encodes each view independently. A shared patch embedding maps the latents to tokens while retaining their view and temporal indices. We sample each arm's action sequence at uniformly spaced temporal indices to match the number of video latent frames, then encode the sampled actions with a shared temporal transformer.

\textbf{Factorized identity encoding.} We factorize token identity into view, role, and actor components, encoded in separate rotary bands alongside the temporal coordinates. We adapt Simplex Rotary Agent Encoding (SRAE) from Gamma-World~\citep{liu2026gammaworld} to encode view identity, while role and actor encodings specify the stream category and its associated actor.

For a pool of $N>1$ view identities, each active view receives a distinct simplex vertex in $d_{\mathrm{id}}\geq N-1$ dimensions. The vertices and rotary phase offsets are
\begin{equation}
 \mathbf s_v=\sqrt{\frac{N}{N-1}}\,\mathbf Q
 \left(\mathbf e_v-\frac{1}{N}\mathbf 1\right),\qquad
 \boldsymbol\phi_v=\alpha\mathbf s_v,\quad v=1,\ldots,N,
\label{eq:wm-simplex}
\end{equation}
where $\mathbf e_v\in\mathbb R^N$ is a one-hot vector and $\mathbf Q$ is a fixed isometric embedding of the zero-sum subspace into $\mathbb R^{d_{\mathrm{id}}}$. The scalar $\alpha>0$ scales the phase offsets. The unit vertices satisfy $\mathbf s_v^\top\mathbf s_{v'}=-1/(N-1)$ for $v\ne v'$, giving equal pairwise distances. For the two wrist views used here, $N=2$ gives opposite simplex vertices, with one phase vector assigned to each wrist-camera stream.

Each coordinate of $\boldsymbol\phi_v$ controls a rotary pair in the view-identity band. Role and actor phases occupy their respective bands, with actor identities determining the action routing in Eq.~\ref{eq:wm-action-conditioning}. Spatial and temporal indices are retained alongside the factorized identities throughout the DiT.

\textbf{Sparse Hub Attention.} Following Gamma-World~\citep{liu2026gammaworld}, we append a small set of learned hub tokens at every latent time. Visual tokens attend only within their own stream and to the hub, whereas hub tokens attend to all wrist streams and other hub tokens. Successive layers therefore exchange cross-view information through a view--hub--view pathway without requiring direct all-to-all visual attention. We additionally condition the hub on the validity-masked mean of the encoded per-arm action states, so the shared representation receives both arms' controls while actor identity preserves view-specific routing. This hub pathway also supports open-world samples whose wrist trajectories remain in separate local coordinate frames.

\subsection{Camera-Aware Modeling}
\label{app:camera}
\textbf{Camera-guided noise warping.} We adopt CameraNoise~\citep{zhao2026cameranoise} to encode relative camera poses into temporally correlated noise. For each view, calibrated intrinsics and consecutive relative camera transforms define geometry-guided reprojection flow on the latent grid using a fixed pseudo-depth. We apply the official noise warping implementation and mix the resulting noise with independent Gaussian noise using variance normalization. Each view uses its own local pose chain, supporting both open-world and in-studio recordings.

\textbf{Geometry-aware conditioning.} Inspired by camera conditioning in PRoPE~\citep{li2025cameras} and Matrix-Game 3.5~\citep{qian2026matrixgame35}, we combine ray-based features with a learned relative projective attention bias. For calibrated views in a shared frame, each ray comprises its camera origin and unit viewing direction. Given intrinsics $\mathbf K_v$, camera-to-world pose $(\mathbf R_t^v,\mathbf p_t^v)$, and homogeneous pixel coordinate $\bar{\mathbf x}$, the six-dimensional ray feature is
\begin{equation}
 \mathbf r_t^v(\bar{\mathbf x})=
 \left[\mathbf p_t^v;\,
 \frac{\mathbf R_t^v\mathbf K_v^{-1}\bar{\mathbf x}}
 {\|\mathbf R_t^v\mathbf K_v^{-1}\bar{\mathbf x}\|_2}\right]
 \in\mathbb R^6.
\label{eq:wm-ray}
\end{equation}
We encode the ray features and aggregate them into hub tokens using geometric affinities. An MLP maps relative camera transforms and intrinsics to per-head attention biases for within-view temporal interactions, while the hub exchanges information across views. These geometry branches operate on samples with valid calibration and shared-frame registration.

\subsection{Training and Distillation}
\label{app:objectives}
\textbf{Progressive teacher training.} We train the bidirectional teacher in three stages using the data pyramid in Section~\ref{sec:data}. Stage~1 uses all three tiers for single-wrist generation with per-arm action conditioning. Stage~2 trains joint dual-wrist generation on open-world and in-studio recordings, introducing CameraNoise and Sparse Hub Attention. Stage~3 uses only calibrated in-studio dual-wrist recordings, adding ray-based conditioning and relative projective attention biases.

\textbf{Training objective.} For clean latents $\mathbf z_0$ and noise $\boldsymbol\epsilon$, we form $\mathbf z_\sigma=(1-\sigma)\mathbf z_0+\sigma\boldsymbol\epsilon$ for $\sigma\in[0,1]$ and minimize the flow-matching objective~\citep{lipman2022flowmatching} over future tokens,
\begin{equation}
 \mathcal L_{\mathrm{FM}}=
 \mathbb E\!\left[
 \left\|v_\theta(\mathbf z_\sigma,\sigma;\mathcal C)
 -(\boldsymbol\epsilon-\mathbf z_0)\right\|_{\mathcal M}^{2}\right],
\label{eq:wm-flow}
\end{equation}
where $\mathcal C=(\mathcal O_{1:H},\mathcal A,\mathcal G)$ collects the conditions in Eq.~\ref{eq:wm-action-conditioning}, and $\|\cdot\|_{\mathcal M}^{2}$ averages over the valid future tokens indexed by $\mathcal M$. Stage~1 uses independent Gaussian noise; Stages~2 and~3 use CameraNoise for samples with valid camera geometry. We keep observation latents clean and optimize the DiT and active conditioning modules with the VAE frozen.

\textbf{Causal adaptation.} We adapt the teacher with block-causal attention, restricting visual and hub tokens to the current and preceding blocks while preserving view--hub connectivity. The action encoder uses causal masking. We train with Eq.~\ref{eq:wm-flow} on a sampled active block, using clean preceding latents as context, to obtain a causal teacher for distillation.

\textbf{Few-step distillation.} We distill the causal teacher into a few-step student by matching denoised endpoints. Both models receive the same initial noise, preceding latents, and action horizon. Let $F_{\theta}^{(K)}$ denote a $K$-step Euler rollout for the active block. We match their denoised endpoints with
\begin{equation}
 \mathcal L_{\mathrm{distill}}=
 \mathbb E\!\left[
 \left\|F_{\theta_{\mathrm S}}^{(K_{\mathrm S})}(\boldsymbol\epsilon;\mathcal C)
 -\operatorname{sg}\!\left[
 F_{\theta_{\mathrm T}}^{(K_{\mathrm T})}(\boldsymbol\epsilon;\mathcal C)\right]
 \right\|_{\mathcal M}^{2}\right],\qquad K_{\mathrm S}<K_{\mathrm T},
\label{eq:wm-distill}
\end{equation}
where $\theta_{\mathrm S}$ and $\theta_{\mathrm T}$ denote the student and frozen teacher parameters, respectively, and $\operatorname{sg}(\cdot)$ denotes the stop-gradient operator. We use $K_{\mathrm T}=30$ and $K_{\mathrm S}=4$, optimizing only the student DiT through its final denoising step. Student-generated preceding blocks provide rollout context during training.

At inference, the Student accepts per-arm actions for the next block and jointly generates the wrist views in four denoising steps. The completed block is appended to a bounded latent history containing both wrist streams, which provides visual context for subsequent predictions. New actions are supplied block by block and routed to their respective wrist streams, while hub tokens maintain cross-view exchange throughout the rollout.

\subsection{BAF Component Errors and Aggregation}
\label{app:idm}
For each arm $i$, the mean component errors over valid evaluation frames are
\begin{equation}
\begin{gathered}
E_{\mathrm{pos}}^i=\frac{1}{|\mathcal V_{\mathrm{pos}}^i|}\sum_{t\in\mathcal V_{\mathrm{pos}}^i}\bigl\|\widehat{\Delta\mathbf p}_t^i-\Delta\mathbf p_t^i\bigr\|_2,\\
E_{\mathrm{rot}}^i=\frac{1}{|\mathcal V_{\mathrm{rot}}^i|}\sum_{t\in\mathcal V_{\mathrm{rot}}^i}\arccos\!\left(\frac{\operatorname{tr}((\widehat{\Delta\mathbf R}_t^i)^\top\Delta\mathbf R_t^i)-1}{2}\right),\\
E_{\mathrm{grip}}^i=\frac{1}{|\mathcal V_{\mathrm{grip}}^i|}\sum_{t\in\mathcal V_{\mathrm{grip}}^i}\bigl|\hat g_t^i-g_t^i\bigr|,
\end{gathered}
\label{eq:idm-component-errors}
\end{equation}
where $(\Delta\mathbf p_t^i,\Delta\mathbf R_t^i,g_t^i)$ is the conditioning action and $(\widehat{\Delta\mathbf p}_t^i,\widehat{\Delta\mathbf R}_t^i,\hat g_t^i)$ is its IDM estimate, with poses relative to the initial frame. Equation~\ref{eq:idm-action-consistency} aggregates the six component errors. The parameters in Section~\ref{sec:idm} define the tolerance bands and decay rate. We average scores over videos with all six errors available. Recovered and conditioning poses use the same coordinate convention and scale; GT videos and baselines lacking full UMI controls use recorded actions as the reference.

\section{Experimental Details}
\label{app:experiments}

\subsection{Baseline Comparisons}
\label{app:baseline-comparisons}
\textbf{Baseline sources.} Text-conditioned baselines are Wan2.2 TI2V-5B~\citep{wan2025wan}, Cosmos Predict2.5-2B~\citep{nvidia2025worldsimulation}, MiniMax-H3~\citep{minimax2026h3,zhao2026minimaxh3reasonphysicalworld}, Cosmos3-Nano-Policy-DROID~\citep{nvidia2026cosmos3}, and RynnWorld-Teleop~\citep{zhao2026rynnworldteleop} (text+image SFT). Action-conditioned baselines are Genie-Envisioner~\citep{liao2025genieenvisioner}, RynnWorld-Teleop (control band), and Gamma-World~\citep{liu2026gammaworld}. Camera-controlled baselines are HY-WorldPlay 5B~\citep{sun2025worldplay}, LingBot-World-Fast~\citep{robbyant2026lingbotworld}, DreamX-World-5B-Cam~\citep{dreamx2026dreamxworld}, Matrix-Game 2.0~\citep{he2025matrixgame2} and 3.0~\citep{wang2026matrixgame3}, YUME 1.5~\citep{mao2025yume15}, and Infinite-World~\citep{wu2026infiniteworld}.

\textbf{Evaluation protocol.} Each valid interval starts at a stride of 189 frames, including the final valid start. Task descriptions are fixed before generation, with missing descriptions obtained from wrist videos using Volcano Engine's LAS service. Frame metrics use the 188 predicted frames at $320\times384$ after the conditioning frame, aligned to reference timestamps. We average over frames, wrists, and seeds within each window, then over windows with equal episode weights.

\begin{figure}[!ht]
\centering
\begin{tikzpicture}
\node[inner sep=0, anchor=south west] (img) {\includegraphics[width=\linewidth]{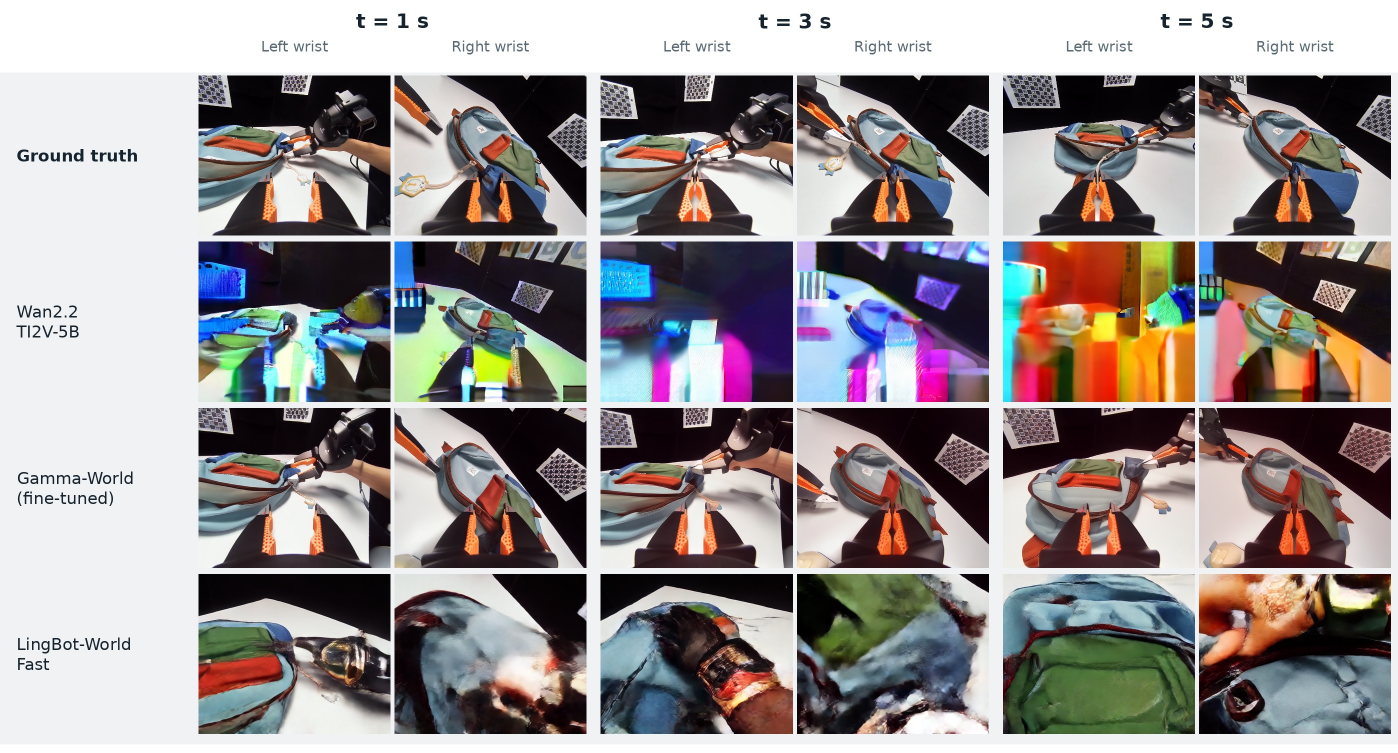}};
\begin{scope}[x={(img.south east)}, y={(img.north west)}]
\tikzset{gripmark/.style={draw=red, line width=0.9pt}, griplabel/.style={fill=white, fill opacity=0.85, text opacity=1, text=red, inner sep=1pt, font=\fontsize{5.5}{6}\selectfont\bfseries, align=center}}
\draw[gripmark] (0.209,0.292) ellipse [x radius=0.035, y radius=0.046];
\draw[gripmark] (0.496,0.292) ellipse [x radius=0.035, y radius=0.046];
\node[griplabel, anchor=north] at (0.492,0.451) {stays open\\(GT: closed)};
\end{scope}
\end{tikzpicture}
\caption{\textbf{Additional temporal comparison.} Additional baselines on the paired-wrist sequence in Figure~\ref{fig:qualitative-comparison}. Red circles mark Gamma-World's gripper staying open as GT closes.}
\label{fig:qualitative-matrix}
\end{figure}

\textbf{Reference and distribution metrics.} FID uses Inception-v3~\citep{szegedy2016inceptionv3} frame features and FVD uses I3D~\citep{carreira2017i3d} video features separately for each wrist and seed. Both distribution metrics use one fixed-duration, fixed-rate, fixed-resolution window per episode. For FVD, pixels are normalized once and resized to $224\times224$ without extractor rescaling.

\textbf{Excluded benchmark metrics.} We also evaluated WorldModelBench~\citep{Li2025WorldModelBench}, iWorld-Bench~\citep{fang2026iworldbench}, and EWMBench~\citep{hu2025ewmbench}, but omit them from the main comparison. WorldModelBench was insensitive to swapped task descriptions on one of two paired windows and returned the same visual-quality answer for original and corrupted videos. iWorld-Bench trajectory scores changed across repeated ViPE~\citep{huang2025vipe} pose estimates and accepted a frozen-frame control, while EWMBench's released gripper tracker produced no valid tracks at its default thresholds and retained large localization errors after lowering them. We therefore retain only the validated VBench and WorldArena dimensions.

\subsection{Bimanual Action Fidelity Analysis}
\label{app:baf-analysis}
\textbf{Evaluation details.} Generated videos use seeds 17 and 29 for IDM inference, with relative-pose supervision and local geometric refinement; the gripper estimator is fixed. Component errors use meters and radians before scoring (Appendix~\ref{app:idm}). Geometry probes and Student sampling use separate evaluation runs from the main comparison.

\textbf{Rendered-control adapters.} Genie-Envisioner uses the Cosmos2-based GE-Sim v0.1 world model. GE-Sim receives projected end-effector markers colored by gripper opening and six-channel camera raymaps, then jointly generates 189-frame wrist videos at 30\,Hz.

\textbf{RynnWorld-Teleop.} For the control entry, we render the mapped UMI pose-and-gripper sequence frame by frame as a synchronized 21-point dual-hand skeleton video, pass it through the Wan2.2 VAE, and add the resulting control features to patch embeddings at scale 0.109; the initial wrist image supplies the visual condition for each view. This route has no numerical action interface.

\textbf{Gamma-World fine-tuning.} We initialize from the existing student checkpoint at iteration 6,400 and reinitialize the action encoders. Each arm supplies window-relative translation, 6D rotation, and gripper opening (10D; Eq.~\ref{eq:per-arm-action}), giving 20D total. We train on paired rectified 189-frame videos at 30\,Hz and $320\times384$ resolution for 3,000 iterations.

\subsection{Applications}
\label{app:applications}
\begin{figure}[!htbp]
\centering
\includegraphics[width=\linewidth]{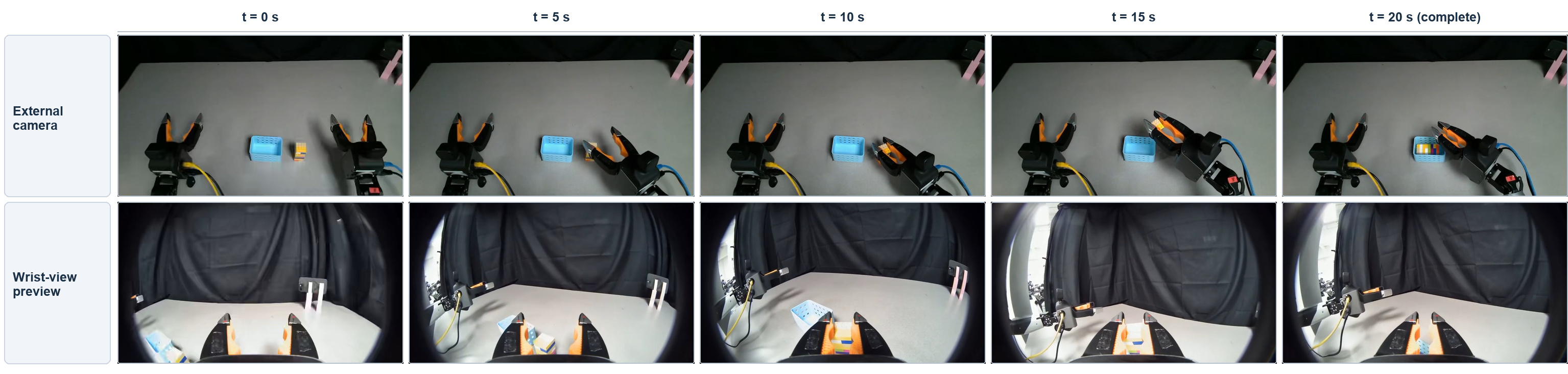}
\caption{\textbf{Physical robot execution and wrist-view preview.} The top row shows external-camera images; the bottom row shows the corresponding wrist-view preview.}
\label{fig:robot-execution}
\end{figure}

\textbf{Physical robot execution.} Figure~\ref{fig:robot-execution} pairs the external-camera view of the physical robot with the corresponding wrist-view preview for the rollout in Section~\ref{sec:exp-realtime}, at five timestamps sampled from initial approach through completed object transport.

\begin{figure}[H]
\centering
\includegraphics[width=0.95\linewidth]{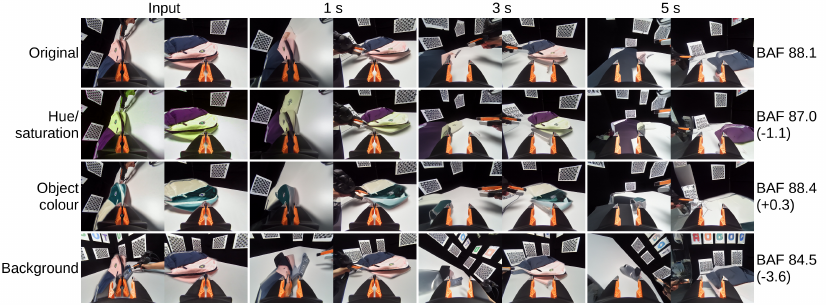}
\vspace{-6pt}
\caption{\textbf{Visual domain randomization.} Rows: original and three initial-frame edits with their BAF; columns: input and generated frames at 1, 3, and 5\,s, each pairing left and right wrists.}
\label{fig:app-domain-rand}
\end{figure}

\textbf{Visual domain randomization.} Figure~\ref{fig:app-domain-rand} applies three edits to both initial frames (hue and saturation shift, object color inversion, background replacement) and regenerates the 189-frame rollout under unchanged controls. BAF (Section~\ref{sec:idm}) screens them: hue and object-color edits stay within one point of the original (87.0 and 88.4 versus 88.1), whereas background replacement lowers BAF by 3.6--6.8 points across three windows and is flagged before policy training.
\vspace{-8pt}

\begin{figure}[H]
\centering
\includegraphics[width=0.95\linewidth]{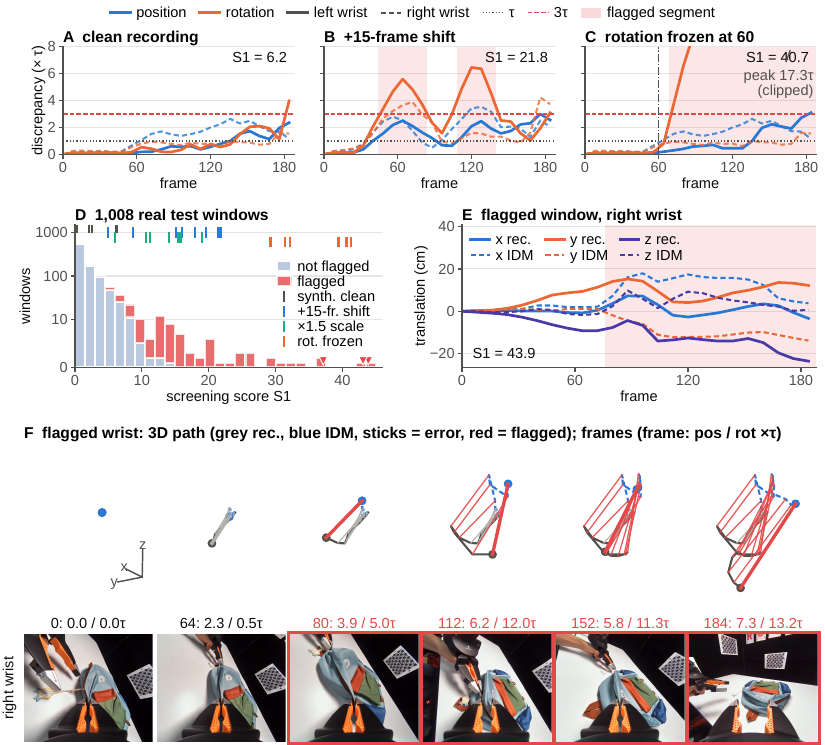}
\vspace{-6pt}
\caption{\textbf{Data quality screening.} (A--C) Per-timestep IDM-versus-recorded discrepancy for a clean recording and two corrupted controls with flagged segments shaded. (D) Screening scores over 1,008 test windows. (E) A flagged real window: recorded versus recovered right-wrist translation. (F) 3D path of the flagged wrist, recorded versus recovered with error sticks in red, and its frames.}
\label{fig:app-screening}
\end{figure}

\textbf{Data quality screening.} Figure~\ref{fig:app-screening} applies the frozen IDM to raw recordings and compares the recovered actions with the recorded SLAM trajectories and gripper controls, normalized by the BAF tolerances of Section~\ref{sec:idm}; a segment is flagged when the discrepancy stays above three tolerances for at least 0.8\,s. Synthetic time-shift, trajectory-scale, and frozen-rotation corruptions are separated from clean recordings with AUROC 0.98, 0.98, and 0.90, with no false alarms on clean recordings.
\vspace{-2pt}

\section{Additional Related Work}
\label{app:related-work}

\textbf{Camera-controlled and multi-view video generation.} Camera control in video diffusion has been extended to object and camera motion~\citep{wang2024motionctrl}, dynamic scenes~\citep{he2025cameractrl2}, analysis of camera conditioning in video transformers~\citep{bahmani2025ac3d}, and 3D-informed generation~\citep{ren2025gen3c}. ReCamMaster~\citep{bai2025recammaster} and TrajectoryCrafter~\citep{yu2025trajectorycrafter} re-render a given video along new camera trajectories, and CamCloneMaster~\citep{luo2025camclonemaster} transfers camera motion from a reference video. SynCamMaster~\citep{bai2025syncammaster} and Collaborative Video Diffusion~\citep{kuang2024cvd} jointly generate several camera views of a shared scene with cross-view consistency. In UMI-WM, each view instead follows its own gripper-mounted trajectory and gripper width, so the two views move independently while observing a shared scene. Warped-noise motion control~\citep{burgert2025gowiththeflow} underlies the camera-guided noise warping adopted in Appendix~\ref{app:camera}. Interactive world models for games~\citep{alonso2024diamond,decart2024oasis,guo2025mineworld,he2025matrixgame2} and general interactive world models such as Astra~\citep{zhu2026astra} similarly generate observations from user controls.

\textbf{Autoregressive and few-step video generation.} UMI-WM builds on the diffusion transformer (DiT) architecture~\citep{peebles2023dit}. Diffusion Forcing~\citep{chen2024diffusionforcing} enables causal rollout with per-frame noise levels. CausVid~\citep{yin2025causvid} converts bidirectional video diffusion models into fast autoregressive generators with distribution matching distillation~\citep{yin2024dmd,yin2024dmd2}, and Self Forcing~\citep{huang2025selfforcing} conditions training on the model's own rollouts. The block-causal adaptation and few-step student in Appendix~\ref{app:objectives} follow this line, matching denoised endpoints (Eq.~\ref{eq:wm-distill}) with student-generated preceding blocks as context.

\textbf{Action-conditioned robot world models.} UniSim~\citep{yang2024unisim} learns an interactive real-world simulator from diverse data, IRASim~\citep{zhu2025irasim} generates robot videos from frame-level trajectories, Cosmos~\citep{nvidia2025cosmos} provides world foundation models for physical AI, and Vid2World~\citep{huang2026vid2world} converts pretrained video diffusion models into interactive world models. Closer to our setting, Ctrl-World~\citep{guo2026ctrlworld} predicts multi-view observations, including wrist views, for policy evaluation and improvement; Mem-World~\citep{zheng2026memworld} adds memory for persistent manipulation, and WristWorld~\citep{qian2025wristworld} generates wrist views from anchor views. These models typically train on large robot datasets such as Open X-Embodiment~\citep{openx2023} and DROID~\citep{khazatsky2024droid}, whereas UMI-WM conditions each wrist view on its own relative camera pose and gripper width.

\textbf{Uses of robot world models.} World models support policy evaluation~\citep{quevedo2026worldgym,li2025worldeval}, bimanual failure detection~\citep{ward2026bimanualfailures}, embodied future prediction~\citep{huang2025enerverse,jiang2025enerverseac,shang2025roboscape}, bimanual data generation~\citep{chen2026craft}, and learning from human videos~\citep{gao2026dreamdojo,chen2026handsonworld}.

\textbf{Joint video--action and latent world models.} Joint video--action models~\citep{li2025uva,zhu2025uwm,shen2025videovla,ye2026dreamzero,zhang2026lingbotva2,yan2026flexpi,egowam2026} predict future observations or world representations together with actions for policy learning. Latent world models instead learn dynamics in latent space for control and planning, as in DayDreamer~\citep{wu2022daydreamer}, TD-MPC2~\citep{hansen2024tdmpc2}, and V-JEPA~2~\citep{assran2025vjepa2}.

\textbf{Actions from video.} Inverse dynamics models label unlabeled video with actions, as in VPT~\citep{baker2022vpt}, while Genie~\citep{bruce2024genie} learns latent actions without labels. DreamGen~\citep{jang2025dreamgen} applies an IDM to world-model rollouts to obtain pseudo-actions for policy training. BAF uses the IDM to compare recovered actions with the conditioning actions.

\textbf{Teleoperation interfaces.} Immersive VR teleoperation systems such as Open-TeleVision~\citep{cheng2024opentelevision} and Bunny-VisionPro~\citep{ding2025bunnyvisionpro} give the operator real-time feedback from the robot, and TelePreview~\citep{guo2024telepreview} previews a virtual arm before commands are executed, whereas the application in Section~\ref{sec:experiments} renders both wrist views from the world model.

\end{document}